# The use of Artificial Intelligence for Intervention and Assessment in Individuals with ASD

Angeliki Sideraki, Christos-Nikolaos Anagnostopoulos

Intelligent Systems Lab, Cultural Technology and Communication, University of the Aegean, Greece
Email: cti24002@ct.aegean.gr, canag@aegean.gr

**Abstract:**

This paper explores the use of Artificial Intelligence (AI) as a tool for diagnosis, assessment, and intervention for individuals with Autism Spectrum Disorder (ASD). It focuses particularly on AI's role in early diagnosis, utilizing advanced machine learning techniques and data analysis. Recent studies demonstrate that deep learning algorithms can identify behavioral patterns through biometric data analysis, video-based interaction assessments, and linguistic feature extraction, providing a more accurate and timely diagnosis compared to traditional methods. Additionally, AI automates diagnostic tools, reducing subjective biases and enabling the development of personalized assessment protocols for ASD monitoring.At the same time, the paper examines AI-powered intervention technologies, emphasizing educational robots and adaptive communication tools. Social robotic assistants, such as NAO and Kaspar, have been shown to enhance social skills in children by offering structured, repetitive interactions that reinforce learning. Furthermore, AI-driven Augmentative and Alternative Communication (AAC) systems allow children with ASD to express themselves more effectively, while machine-learning chatbots provide language development support through personalized responses.The study presents research findings supporting the effectiveness of these AI applications while addressing challenges such as long-term evaluation and customization to individual needs. In conclusion, the paper highlights the significance of AI as an innovative tool in ASD diagnosis and intervention, advocating for further research to assess its long-term impact.

## 1. Introduction

### 1.1. Problem statement (indicative)

The increasing prevalence of Autism Spectrum Disorder (ASD) has highlighted the urgent need for accurate diagnosis and effective intervention strategies. Traditional diagnostic methods rely on behavioral observations and standardized assessments, which can be subjective, time-consuming, and dependent on the expertise of professionals. As a result, many children with ASD experience delayed diagnosis, limiting their access to early intervention programs that are crucial for their development and social integration.

At the same time, existing intervention approaches often lack personalization, making it challenging to address the diverse needs of individuals with ASD. Many conventional therapeutic methods depend heavily on human interaction, which may not always be engaging or effective for children with social and communication difficulties. The demand for scalable, adaptive, and evidence-based solutions has therefore increased, pushing the field towards technology-assisted interventions.

In this context, Artificial Intelligence (AI) has emerged as a promising tool for both diagnosis and intervention in ASD. AI-powered systems, including machine learning algorithms, deep learning models, and social robots, have demonstrated potential in identifying behavioral patterns, speech irregularities, and nonverbal cues, leading to faster and more objective assessments. Additionally, AI-driven educational robots, adaptive communication systems (AAC), and personalized chatbot assistants offer new possibilities for enhancing social skills, communication, and learning experiences for individuals with ASD.

Despite these advancements, significant challenges remain. The reliability, ethical considerations, and long-term effectiveness of AI-based ASD interventions require further investigation. Moreover, the integration of AI technologies into clinical and educational settings demands rigorous validation, user-friendly designs, and appropriate training for educators, therapists, and caregivers.This study aims to explore the current applications, benefits, and challenges of AI in ASD diagnosis and intervention. By examining existing research and real-world implementations, it seeks to determine how AI can be leveraged to create more accessible, efficient, and individualized solutions for individuals with ASD, ultimately improving their quality of life and developmental outcomes.

### 1.2. Proposed solution – objectives

To address the challenges associated with Autism Spectrum Disorder (ASD) diagnosis and intervention, this study proposes the integration of Artificial Intelligence (AI) as a tool for early detection, personalized intervention, and long-term monitoring. By leveraging AI-driven technologies such as machine learning, deep learning, and social robotics, the aim is to improve the accuracy, efficiency, and accessibility of ASD-related assessments and therapies.

The proposed solution focuses on enhancing diagnostic accuracy by using AI models that analyze biometric, linguistic, and behavioral data to detect ASD-related patterns more objectively than traditional methods. Additionally, AI-powered tools, such as computer vision and speech analysis, can automate and streamline assessment procedures, reducing the time required for evaluations while minimizing human bias. Personalized intervention strategies can be developed through AI-driven adaptive learning systems and social robots, which tailor therapy sessions based on the individual needs of each child. Furthermore, AI-enhanced augmentative and alternative communication (AAC) tools can assist non-verbal individuals in expressing their thoughts and emotions, improving their ability to communicate. The integration of AI-based wearable devices and smart applications also enables continuous monitoring of progress and behavioral changes over time, providing real-time insights to caregivers and professionals.

This research aims to evaluate the effectiveness of AI in ASD diagnosis by comparing its performance with conventional assessment methods and investigating its role in personalized interventions. Special focus will be given to AI-based tools such as educational robots, AAC systems, and chatbot-assisted communication. The study will also identify ethical and practical challenges in AI implementation within clinical, educational, and home settings while exploring the long-term impact of AI-based interventions on sustainability, engagement levels, and developmental outcomes. Finally, it seeks to provide recommendations for integrating AI tools into multidisciplinary ASD support frameworks, fostering collaboration between therapists, educators, and caregivers. By achieving these objectives, this study aspires to contribute to the advancement of AI-driven solutions that improve the quality of life and developmental potential of individuals with ASD.

## 2. AI and Intervention in ASD

Artificial intelligence (AI) has emerged as a powerful ally in the intervention and support of children with autism, offering personalized and dynamic solutions that can significantly improve communication, social interaction, and the learning process. AI applications in autism intervention include educational robots, virtual reality (VR), smart communication applications, and customized learning environments, allowing children to develop skills at their own pace and in their own way (Sideraki & Drigas, 2021).

Social robots are among the most successful applications of AI in autism intervention. Robots such as NAO and Kaspar are designed to help children with autism develop essential social skills, such as eye contact, emotional understanding, and participation in dialogues. Studies show that children with autism often prefer interacting with robots over humans, as robots provide more predictable and structured responses, reducing the anxiety and confusion that human communication can sometimes cause (Sideraki & Drigas, 2021). Kaspar, for instance, is used to teach children the importance of non-verbal communication, while NAO helps improve verbal expression and emotional recognition (Nan Xing, 2024).

Communication difficulties are one of the biggest challenges for children with autism. AI-based Augmentative and Alternative Communication (AAC) systems are designed to help children express their needs, thoughts, and emotions. These devices use advanced technologies, such as word prediction and voice recognition, to facilitate communication and provide a more natural experience. AAC tools can be adapted to the needs of each child, promoting autonomy and reducing the frustration that often accompanies communication difficulties (Nan Xing, 2024).

The education of children with autism can be significantly enhanced through AI-based educational applications and programs. Applications such as Social Mind Autism and ABA Genie offer personalized education that considers each child's individual needs and abilities. Using techniques such as behavioral analysis and continuous feedback, these systems adapt educational content to maintain engagement and enhance the learning process (Sideraki & Drigas, 2021).

Furthermore, the use of AI technologies enables educators and therapists to collect and analyze data on children's progress, allowing them to adjust intervention strategies in real time. This dynamic approach improves the effectiveness of interventions and contributes to maximizing the learning experience for children with autism (Nan Xing, 2024).

**2.1 Chatbot Applications in Autism Support**

According to the study by Li et al. (2020), the development of a deep learning-based chatbot specifically designed for children with autism spectrum disorder aims to improve their social and communication skills. The chatbot was trained using a dataset of 1.7 million child conversation question-answer pairs, which was cleaned and filtered to retain 400,000 suitable sentences for model training. A generative model combining Bi-LSTM and attention mechanisms was implemented to generate natural and semantically coherent responses. This approach enhanced the chatbot's ability to respond to single-word inputs and adapt to the linguistic characteristics of children with autism. The chatbot's quality was evaluated through comparisons with two other chatbots, assessing anthropomorphism, emotional response, accessibility, and overall user impression. Results indicated that the chatbot exhibited a higher understanding of user intent, used polite language, and displayed a positive personality—key features for children with autism. Despite the promising results, the study highlighted the need for further clinical trials to assess the chatbot's therapeutic effectiveness in real intervention settings (Li et al., 2020).

Subsequently, the systematic review by Federici et al. (2020) focused on the perceived quality of interaction between chatbots and individuals with disabilities or special needs to identify evaluation methodologies and existing gaps in the literature. Out of the 192 articles analyzed, only 15 included assessments from actual users with special needs, underscoring the limited research focus on user experience. Most studies primarily assessed the clinical effectiveness of chatbots without considering critical factors such as usability, satisfaction, and integration into the daily lives of individuals with autism. The review found that no universally accepted framework exists for evaluating interaction quality, making it difficult to compare different chatbots and obtain objective data on their impact on users. The researchers emphasized the need for standardized evaluation methods incorporating factors such as accessibility, user acceptance, and actual improvements in communication and learning. This study highlights the gap between technological development and the assessment of real user experience, pointing out that the lack of clear evaluation criteria creates barriers to adopting chatbots as support tools for individuals with autism (Federici et al., 2020).

Additionally, according to the study by Ireland et al. (2018), chatbot use in educational settings for children with autism was investigated to determine their ability to tailor responses to students' individual needs. The chatbot was evaluated through interactive scenarios in classroom environments, analyzing factors such as children's comprehension of responses, their ability to sustain conversation, and their overall acceptance of the chatbot. Educators involved in the study noted that the chatbot provided a safe and predictable communication environment, encouraging children to engage more than in traditional social settings. Findings indicated that the chatbot contributed to linguistic development by offering personalized responses that matched each student's comprehension level. However, the study also highlighted challenges, including difficulties in adapting the chatbot to children with severe language impairments and the need for further refinement of dialogue management mechanisms to ensure coherence and appropriateness of responses (Ireland et al., 2018).

Comparing these studies reveals critical issues related to chatbot development and evaluation for children with autism. On one hand, advancements in deep learning and natural language processing enable the creation of more adaptive and intelligent systems that can provide valuable support in therapeutic and educational settings (Li et al., 2020; Ireland et al., 2018). On the other hand, the absence of a standardized evaluation framework and the limited research on user experience make it challenging to objectively assess the effectiveness of these systems (Federici et al., 2020). Future research should focus on developing universally accepted evaluation methodologies that consider not only chatbot functionality but also their impact on enhancing communication and social integration for children with autism.

**2.2 Social Robots for Autism Intervention**

According to the study by Di Nuovo et al. (2020), the use of social robots in interventions for children with Autism Spectrum Disorder (ASD) presents significant advantages, as robots can provide repetitive, predictable, and non-judgmental interactions that help children develop social skills . The main objective of this study is to evaluate the effectiveness of robots in this context by examining children's interactions with robots and the behavioral changes observed during the sessions.

In the introduction, the authors present the theoretical framework of using social robots in special education and analyze previous studies that have focused on similar applications. Specifically, they highlight that social robots can be adapted to each child's needs through personalized interventions, making them particularly useful for children with ASD.To conduct the study, the researchers used an experimental framework involving children diagnosed with ASD, aged 5-10 years . The protocol included multiple interaction sessions with the social robot NAO, which has been extensively used in related studies due to its flexibility in communication and movement. During the sessions, children engaged in various activities guided by the robot, such as imitation games, facial expression recognition, and social conversations. The children's behavior was assessed through direct observation, questionnaires administered to parents and teachers, and recordings of interactions via video analysis (Di Nuovo, et al., 2020).

The study results indicated that using social robots can significantly improve social interaction skills in children with ASD, particularly in areas such as maintaining eye contact, verbal communication, and emotion recognition.. Additionally, increased engagement in activities and a reduction in avoidance behaviors were observed during the sessions.However, the study also highlights certain limitations, such as the need to tailor robots to each child's individual needs and the potential decline in children's interest after prolonged use. The authors suggest further research on larger sample sizes and an investigation into the long-term effectiveness of social robots in autism interventions (Kim et al., 2020). Overall, the study confirms that social robots are a promising tool for improving social interaction in children with ASD, offering significant advantages but also presenting challenges that need to be addressed in future research (Di Nuovo, et al., 2020).

According to the study by Rakhymbayeva et al. (2021), social robots have the potential to significantly enhance the engagement and emotional involvement of children with Autism Spectrum Disorder (ASD) through long-term interventions. The aim of the study was to investigate whether prolonged interaction with a social robot could increase the children's attention span and enhance their emotional responses (Rakhymbayeva et al., 2021).

To achieve this goal, the researchers designed an experimental protocol that involved several weeks of robot interaction sessions. The participants were divided into two groups: an experimental group, where children engaged in sessions with the social robot, and a control group, where children participated in typical educational activities without the robot. The sessions were videotaped, and the researchers analyzed the children's behavior, focusing on their attention towards the robot, their emotional reactions, and the duration of their involvement in each activity (Rakhymbayeva et al., 2021).

The results showed that children interacting with the social robot exhibited significant improvements in maintaining attention. It was observed that their attention span gradually increased, with children staying focused on the robot activities for longer periods compared to the control group (Rakhymbayeva et al., 2021). Additionally, their emotional involvement was more intense, as they expressed more frequent signs of joy, interest, and social responsiveness during the sessions.The researchers concluded that continuous interaction with social robots could enhance the emotional development of children with autism, creating an environment that promotes their social integration (Rakhymbayeva et al., 2021). They emphasized that the use of social robots as an intervention tool should be individualized and tailored to the needs of each child to achieve maximum benefit.

Furthermore, the study by Fachantidis et al. (2020), focused on the use of adaptive social robots in supporting children with autism, with particular emphasis on the robot's behavior customization during interaction. The aim of the research was to determine whether the adaptive behavior of the robot could positively influence the children's attention span and engagement in activities (Coninx et al., 2016).To achieve this goal, the researchers designed a dynamic interaction system in which social robots could adjust their behavior in real time based on the children's reactions. The system incorporated sensors that detected the emotional state and posture of the child, allowing the robot to modify its expressiveness, movements, and communication style accordingly (Fachantidis et al., 2020).

The study included interaction sessions where children participated in activities with the robot, which adapted to their needs by analyzing their behavior. Data were collected through videotaped sessions and assessment of the children's attention and participation .The results showed that children interacting with the adaptive robot maintained attention for a longer period compared to previous non-adaptive robot versions. The robot's ability to adjust its reactions based on the child's behavior led to increased involvement in the activities and greater participation in social interactions (Fachantidis, et al., 2020).

Additionally, it was observed that the use of the adaptive robot enhanced the children's non-verbal communication forms, such as maintaining eye contact and using gestures to express their needs (Coninx et al., 2016). The researchers concluded that the adaptability of social robots is a critical factor for the success of interventions and suggested that future research should focus on further improving the adaptation algorithms to enhance the personalization of interventions for each child (Fachantidis, et al., 2020).

## 2.3 NAO Robot

According to the study by Baraka et al. (2019), the use of social robots in interventions for children with Autism Spectrum Disorder (ASD) presents significant advantages, as robots can provide repetitive, predictable, and non-judgmental interactions that help children develop social skills. The main objective of this study is to evaluate the effectiveness of robots in this context by examining children's interactions with robots and the behavioral changes observed during the sessions.In the introduction, the authors present the theoretical framework of using social robots in special education and analyze previous studies that have focused on similar applications. Specifically, they highlight that social robots can be adapted to each child's needs through personalized interventions, making them particularly useful for children with ASD.

To conduct the study, the researchers used an experimental framework involving children diagnosed with ASD, aged 5-10 years (Miller et al., 2023). The protocol included multiple interaction sessions with the social robot NAO, which has been extensively used in related studies due to its flexibility in communication and movement.During the sessions, children engaged in various activities guided by the robot, such as imitation games, facial expression recognition, and social conversations. The children's behavior was assessed through direct observation, questionnaires administered to parents and teachers, and recordings of interactions via video analysis (Baraka, et al., 2019). The study results indicated that using social robots can significantly improve social interaction skills in children with ASD, particularly in areas such as maintaining eye contact, verbal communication, and emotion recognition. Additionally, increased engagement in activities and a reduction in avoidance behaviors were observed during the sessions (Baraka, et al., 2019).

However, the study also highlights certain limitations, such as the need to tailor robots to each child's individual needs and the potential decline in children's interest after prolonged use. The authors suggest further research on larger sample sizes and an investigation into the long-term effectiveness of social robots in autism interventions.Overall, the study confirms that social robots are a promising tool for improving social interaction in children with ASD, offering significant advantages but also presenting challenges that need to be addressed in future research (Baraka, et al., 2019).

According to the study by Arsić and colleagues (2022), the use of social robots is based on their ability to provide structured, repetitive, and predictable interactions, which is considered crucial for children with Autism Spectrum Disorder (ASD). The main goal of this study is to assess the extent to which the interaction of children with ASD with social robots can lead to improvements in their social skills. The authors examine both the technical capabilities of the robots and the responses of the children, contributing to the broader discussion on the utility of robotic technology in special education (Arsić, et al., 2022).

In the introduction of the article, the theoretical background of the use of social robots in educational and therapeutic interventions is analyzed. Particular emphasis is given to the characteristics of children with ASD, who often struggle with the development of social skills and require approaches that facilitate their communication with the environment. The authors point out that while traditional intervention methods have proven effective, the integration of technology can enhance the learning process and offer personalized solutions (Arsić, et al., 2022).For the study, the researchers used an experimental design that involved children with diagnosed ASD aged 5 to 10 years. The social robot chosen for the research was NAO, which has been extensively used in related studies due to its flexibility in communication and ability to adapt to the needs of the children (Kim et al., 2020).

The interaction sessions were organized in a structured environment and included activities aimed at enhancing the social skills of the children. During these sessions, the children participated in imitation games, emotion recognition tasks, and social communication activities with the robot, which responded with vocal commands and movements. The children's assessment was conducted through direct observation, questionnaires completed by parents and teachers, and analysis of video-recorded interactions (Arsić, et al., 2022).The researchers adopted both quantitative and qualitative analysis methods. Quantitative data was derived from comparisons before and after the intervention using statistical tools such as ANOVA to measure changes in the children's behavior. The qualitative analysis was based on interviews with parents and teachers to better understand the children's experience and their perception of the usefulness of social robots (Arsić, et al., 2022).

The results of the research showed that interaction with the NAO robot significantly contributed to the improvement of specific social skills in children with ASD. Specifically, there was an increase in eye contact, verbal communication, and emotion recognition abilities. The children who participated in the sessions showed greater engagement in the activities and a reduced tendency to avoid social interaction, indicating the positive impact of social robots on social development (Arsić, et al., 2022).

However, the researchers identified some limitations that should be considered. First, the need for adaptation of the robots to the individual needs of each child is deemed crucial, as a general approach may not be equally effective for all children with ASD (Arsić, et al., 2022). Secondly, it was found that in some cases, the children's interest in the sessions decreased after a certain period, suggesting the need for further differentiation of activities and enrichment of the interactions (Kim et al., 2020).

The authors conclude that social robots can be a particularly useful tool for autism interventions, offering an innovative and personalized approach to the education and therapy of children with ASD. They recommend continuing research with larger sample sizes and exploring the long-term effectiveness of social robots in educational and therapeutic settings. Additionally, they suggest the future integration of artificial intelligence so that robots can adapt more dynamically to the needs of each child, thus enhancing their educational value (Arsić, et al., 2022).

The study confirms that social robots have the potential to significantly improve the social interaction of children with ASD, contributing to the development of skills that are crucial for their social integration. While there are challenges to be addressed, the findings suggest that robots can function as complementary tools to existing interventions, enhancing educational and therapeutic approaches for children with autism (Arsić, et al., 2022).

Finnaly, according to the study by Urdanivia Alarcon et al. (2021), the use of robot-based video interventions for children with Autism Spectrum Disorder (ASD) can be an effective alternative to contemporary remote education. The main objective of the study was to investigate the effectiveness of using robot-based videos, specifically the NAO robot, in facilitating the educational process for children with ASD. The researchers aimed to understand whether robotic videos could enhance the attention, participation, and learning experience of these children (Urdanivia Alarcon et al., 2021).

The study's methodology was based on an experimental design in which robotic video interventions were used for children with ASD. Participants were divided into two groups: a control group that followed conventional remote education and an experimental group that was exposed to videos featuring robotic characters. The researchers recorded and analyzed the children's behavior using quantitative and qualitative measures (Urdanivia Alarcon et al., 2021).

The study's results showed that children in the experimental group exhibited higher levels of attention and participation compared to the control group. Additionally, an improvement in communication skills and interaction with lesson content was observed. The researchers found that using robotic videos can be a valuable tool for educating children with ASD, providing a more engaging and interactive learning environment (Urdanivia Alarcon et al., 2021). The study concluded that robotic video interventions represent a promising strategy for improving remote education for children with ASD. Further investigation of the topic through long-term studies was suggested, along with the integration of this technology into educational programs (Urdanivia Alarcon et al., 2021).

**2.4 Kaspar Robot**

According to the study by Mengoni et al. (2017), the research focuses on the use of the social robot Kaspar as a support tool for children with Autism Spectrum Disorder (ASD) to improve their social skills. The authors highlight that social interaction is one of the most significant challenges faced by children with ASD, and therefore, the introduction of robotic systems can serve as a bridge for developing their communication abilities. The methodology of the study was based on qualitative and quantitative analysis of the interaction between children with ASD and Kaspar. The sessions were conducted in a controlled environment, where the robot was used to simulate various facial expressions and social behaviors, such as hand-waving or maintaining eye contact. The children's responses were recorded through video recordings and analyzed using behavioral observation tools. Additionally, data were collected from parents and therapists to assess the intervention's impact on real social situations (Mengoni et al., 2017).

The study's findings showed that Kaspar's presence enhanced children's engagement in social interactions. The children demonstrated increased attention toward the robot as well as greater willingness to communicate through verbal and non-verbal cues. Furthermore, it was observed that the skills acquired during the sessions were partially transferred to interactions with adults and peers. However, the impact of the intervention varied depending on the individual profile of each child, indicating the need for personalized interventions and longer training durations (Mengoni et al., 2017).

In addition, the study by Karakosta et al. (2019), investigates the effect of interacting with Kaspar on the development of visual perspective-taking (VPT) skills in children with ASD. The ability to understand others' visual perspectives is considered crucial for social development and is a common challenge for children with ASD. The authors argue that using a social robot can help children develop this skill through repeated experiences and interactions (Karakosta et al., 2019).

The research methodology included experimental intervention sessions, involving thirteen children with ASD from a special education school. During the sessions, Kaspar performed various movements and expressions, while the children were asked to determine what the robot could see from its position. The children's performance was assessed both before and after the intervention, using quantitative measurements and qualitative analyses of their behavioral responses (Karakosta, et al., 2019).

The results indicated that children showed significant improvement in their ability to recognize Kaspar's perspective. Most participants were able to accurately predict what the robot "saw" in different situations, suggesting that the intervention had a positive effect on theory of mind development. However, the authors note that the findings were limited due to the small sample size, and further studies are needed to assess the long-term effectiveness of the intervention. Additionally, they highlight the need for more personalized approaches, as some children did not exhibit the same level of progress as others (Karakosta, et al., 2019).

Finally, in the study by Lakatos et al. (2021), the authors focus on whether interaction with Kaspar can lead to the generalization of social skills from intervention sessions to real-life social situations. The researchers point out that while robots have been proven effective in improving specific skills, transferring these skills to everyday life remains one of the greatest challenges in ASD interventions (Wooldridge et al., 2021).The study was conducted in a special education school, where structured intervention sessions were carried out with children of various ages and cognitive levels. The researchers recorded children's behavior during the sessions and later observed whether the acquired skills also appeared in other social settings, such as in the classroom or playground. Additionally, interviews were conducted with parents and teachers to examine whether the improvements observed in the intervention sessions persisted in natural environments (Lakatos, et al., 2021).

The study's findings indicated that children demonstrated increased social engagement during the sessions with Kaspar, including eye contact, imitation, and verbal communication. Moreover, parents and teachers reported improvements in children's social skills, although the generalization of these skills to everyday life was not consistent for all children. The authors argue that the effectiveness of the intervention depends on factors such as the duration of exposure to the robot, the individual profile of each child, and the support (Lakatos, et al., 2021).

### 3. Diagnosis of autism through AI

The diagnosis of autism is a multifaceted process that combines behavioral evaluations with advanced technological approaches, including neuroimaging and artificial intelligence. Studies suggest that one of the most consistent neuroimaging findings in autism involves alterations in the Default Mode Network (DMN), a system of interconnected brain regions involved in self-referential thinking, memory retrieval, and social cognition. Research has shown that autistic individuals exhibit reduced connectivity between DMN subsystems, particularly between the medial prefrontal cortex and the posterior cingulate cortex, which are crucial for social processing and mentalizing. These differences indicate a developmental delay in the functional organization of the brain, as neurotypical individuals tend to show increased modularization of the DMN with age, whereas autistic individuals experience a less pronounced shift (Bathelt & Geurts).

In addition to functional connectivity differences, structural abnormalities in the brain have been identified through deep learning applications in neuroimaging analysis. Machine learning models trained on large-scale MRI datasets have highlighted atypicalities in subcortical structures, including the basal ganglia, which plays a role in motor control, learning, and behavior regulation. These findings align with clinical observations that individuals with autism often present with repetitive behaviors, restricted interests, and difficulties in cognitive flexibility, which may be linked to disruptions in these neural circuits. Such AI-driven models are not only advancing our understanding of the neurological underpinnings of autism but are also improving diagnostic accuracy by identifying key biomarkers that distinguish autistic from neurotypical brain patterns (Ke et al.).

Beyond neuroimaging, artificial intelligence has been leveraged to enhance and accelerate behavioral diagnosis. Traditional diagnostic tools, such as the Autism Diagnostic Interview-Revised (ADI-R), involve lengthy assessments that require trained professionals and extensive parental interviews, often taking hours to complete. However, machine learning algorithms have demonstrated that a significantly reduced subset of questions—focusing on key aspects of social communication, eye contact, play behavior, and developmental milestones—can predict autism with near-perfect accuracy. By analyzing large datasets of ADI-R responses, researchers have identified just seven critical questions that effectively classify individuals on the autism spectrum while maintaining the reliability of the full diagnostic assessment. This innovation has the potential to streamline the diagnostic process, enabling earlier identification and intervention, which is critical for improving developmental outcomes (Wall et al.).

The integration of neuroimaging, deep learning, and artificial intelligence in autism research is transforming the landscape of diagnosis, making it more precise, efficient, and accessible. These advancements not only aid in early detection but also contribute to a deeper understanding of the neurological mechanisms underlying autism, paving the way for more targeted therapeutic strategies. As research progresses, the combination of computational tools and behavioral assessments will continue to refine the diagnostic process, ultimately improving the quality of life for individuals on the autism spectrum and their families (Florio et al., 2009).

**3.1 Autism Diagnostic Interview-Revised (ADI-R) through AI**

The purpose of Bone et al. (2016), study is to investigate the potential for improving autism diagnostic tools through the use of machine learning. Existing methods for diagnosing autism rely on clinical evaluations and structured interviews, which, although effective, present limitations in terms of time and the required expertise of clinical evaluators. The authors aim to develop an artificial intelligence model that can integrate data from multiple diagnostic tests, improving both the efficiency and accuracy of the screening process. Also, the methodology involves the use of existing diagnostic tools such as the Autism Diagnostic Observation Schedule (ADOS) and the Autism Diagnostic Interview-Revised (ADI-R), combined with machine learning techniques. The researchers analyzed data from a large number of participants, applying algorithms that allow for more accurate prediction of autism. The data came from clinical evaluations and automated behavioral analyses, which were categorized using advanced computational methods (Bone et al., 2016).

The results show that machine learning techniques can significantly increase the accuracy of autism diagnostic tools while reducing the time and need for human intervention. The proposed model achieved higher predictive accuracy than traditional assessment tools, demonstrating the potential for integrating artificial intelligence into the diagnostic process. Furthermore, the combined use of multiple tools led to better detection of individuals with autism compared to the use of a single diagnostic method (Bone et al., 2016). The authors conclude that incorporating such technologies can improve autism diagnosis, enabling earlier and more reliable detection of the disorder, while also reducing costs and the need for extensive clinical evaluations.

The article by Wall et al. (2012), deals with the use of machine learning techniques to improve the autism diagnostic process, focusing on shortening the time required for evaluation through the Autism Diagnostic Interview-Revised (ADI-R). Their goal is to develop an AI model that can predict autism diagnosis using only a limited set of questions from the ADI-R, aiming to reduce the time and resources needed for clinical diagnosis.

The methodology includes analyzing data from the Autism Genetic Research Exchange (AGRE), where the authors used machine learning algorithms to detect autism based on responses to 93 questions. The study revealed that using only 7 of these questions, it was possible to achieve a diagnostic accuracy of 99.9% (Wall et al., 2012). The authors performed further tests with two independent data sets (Simons Foundation and Boston Autism Consortium) and found that the accuracy remained nearly 100%, making the model useful for rapid and accurate diagnosis. The results suggest that applying AI to autism detection can improve the efficiency of the diagnostic process, saving time and resources, while maintaining diagnostic accuracy (Wall et al., 2012).

Additionally, the study by Song et al. (2019), examines the use of artificial intelligence to improve autism diagnosis, focusing on the analysis of behavioral and communicative patterns through machine learning. The research team concentrated on analyzing large datasets from diagnostic tools and children's behaviors, aiming to understand how the diagnostic process can be accelerated and how individuals with autism can be identified more quickly. The methodology involves using data from the ADI-R and other clinical studies, which are processed with AI algorithms to develop an accurate and efficient analysis system. Their goal is to create tools that can diagnose autism more quickly and accurately, without the need for extensive and time-consuming interviews (Song et al., 2019).

The results show that AI can offer faster detection and diagnosis, reducing the time required for assessing children with autism. This allows for quicker intervention and better outcomes for individuals with autism, while also reducing the need for human intervention and minimizing errors in diagnosis (Song et al., 2019).

According to Choi et al. (2020), the application of artificial intelligence (AI) has the potential to improve diagnostic classification in autism spectrum disorder (ASD). The primary goal of their study was to explore the use of machine learning (ML) techniques in facilitating early and accurate ASD diagnosis in Korean patients. Given the complex nature of ASD, which lacks a definitive biomarker, the researchers aimed to assess whether AI-driven approaches could enhance diagnostic precision and efficiency (Choi et al., 2020).

The study employed AI algorithms to analyze a dataset of Korean patients diagnosed with ASD. The methodology included data preprocessing, feature selection, and classification using various machine learning models. Researchers utilized diagnostic tools such as the Autism Diagnostic Observation Schedule (ADOS) and the Autism Diagnostic Interview-Revised (ADI-R) to provide reliable input for the AI system. The models were evaluated based on their accuracy, sensitivity, and specificity in distinguishing ASD from other developmental disorders (Choi et al., 2020).

The findings indicated that AI-based classification models could significantly improve diagnostic accuracy for ASD. Certain machine learning algorithms outperformed traditional diagnostic approaches, demonstrating high sensitivity and specificity. Additionally, AI-assisted analysis enabled faster decision-making, reducing the time required for clinical assessments. The study highlighted the potential for integrating AI into psychiatric diagnostics to support clinicians in making more objective and data-driven decisions (Choi et al., 2020). AlsomChoi et al. (2020) concluded that AI applications hold promise in refining ASD diagnostic methods, particularly in populations where access to specialized psychiatric expertise is limited. The authors recommended further research to optimize machine learning models and integrate them into clinical practice. They also emphasized the need for larger, more diverse datasets to enhance generalizability and reliability (Choi et al., 2020).

### 3.2 Machine Learning (ML)

The study by Uddin et al. (2023) focuses on developing an integrated machine learning framework that leverages both statistical techniques and clinical data to improve the detection of autism spectrum disorder (ASD). The researchers highlight that current diagnostic methods are primarily behavioral, making them susceptible to subjective errors and delays in intervention (Uddin et al., 2023). The methodology includes collecting data from ASD patients and control groups, using machine learning algorithms such as Support Vector Machines (SVM), Decision Trees, and Deep Neural Networks (DNN), and performing cross-validation of results. Additionally, data preprocessing is conducted to enhance model performance (Uddin et al., 2023).

The results indicate that the proposed machine learning framework achieves exceptionally high accuracy in ASD diagnosis, with deep neural networks performing the best, exceeding 90% accuracy (Uddin et al., 2023). The study underscores that behavioral and cognitive markers are particularly useful in distinguishing individuals with autism from control groups. However, the authors emphasize the need for further validation of the models on larger datasets (Uddin et al., 2023).

The research by Voinsky et al. (2023), investigates the use of blood RNA analysis as a biological marker for ASD diagnosis, relying on machine learning techniques. The researchers argue that genetic and molecular markers can provide an objective and reliable approach that reduces dependency on behavioral assessments (Voinsky et al., 2023).The study involves collecting blood samples from children with ASD and control groups, analyzing RNA gene expression profiles using bioinformatics techniques, and employing algorithms such as Random Forest, linear regression, and deep learning. The analyses focus on identifying gene expression patterns associated with ASD (Voinsky et al., 2023).

The results show that certain blood RNA profiles serve as strong biomarkers for ASD diagnosis, with machine learning models achieving over 85% accuracy (Voinsky et al., 2023). The authors note that this method could complement existing diagnostic approaches and emphasize the need for further research on larger samples (Voinsky et al., 2023).

The study by Bone et al. (2016), examines how machine learning can improve the effectiveness of diagnostic tools for autism. The researchers focus on using machine learning algorithms to enhance the accuracy and consistency of tools such as the Autism Diagnostic Observation Schedule (ADOS) and the Social Responsiveness Scale (SRS) (Bone et al., 2016).The methodology involves processing data from existing diagnostic assessments and applying machine learning techniques to improve ASD classification. The researchers analyze how multidimensional data processing can contribute to more precise diagnoses (Bone et al., 2016).

The results indicate that incorporating machine learning into traditional diagnostic tools significantly enhances the sensitivity and specificity of predictions. Specifically, combined models utilizing multiple data sources provide more reliable results than single diagnostic methods (Bone et al., 2016). The authors highlight that applying these techniques can reduce false positive and false negative diagnoses, leading to earlier and more accurate ASD detection (Bone et al., 2016).

This analysis demonstrates the contribution of machine learning to improving ASD diagnosis. The studies suggest that artificial intelligence can be utilized for the development of novel biological diagnostic methods, such as RNA analysis (Voinsky et al., 2023), or for improving existing diagnostic tools (Bone et al., 2016). Additionally, the combined use of data and machine learning techniques can provide more accurate and reliable ASD diagnostic results (Uddin et al., 2023). However, further studies are needed to validate these findings and translate them into clinical applications (Bone et al., 2016; Uddin et al., 2023; Voinsky et al., 2023).

In addition, Kumar and Das (2021), explored the application of multiple machine learning (ML) techniques in diagnosing autism spectrum disorder (ASD). The study aimed to evaluate the effectiveness of different ML models in accurately classifying ASD cases. Given the increasing prevalence of ASD and the challenges in early diagnosis due to the absence of definitive biomarkers, the authors sought to identify the most reliable ML approaches for diagnostic purposes (Kumar & Das, 2021). The study utilized a dataset of ASD diagnostic cases and applied various ML techniques, including decision trees, support vector machines (SVM), and deep learning models. The methodology involved data preprocessing, feature extraction, and model training using supervised learning algorithms. The performance of each model was assessed based on accuracy, precision, recall, and F1-score (Kumar & Das, 2021).

The results demonstrated that ML models could significantly improve ASD diagnosis by identifying patterns within behavioral and clinical data. Among the tested models, deep learning techniques achieved the highest classification accuracy. Additionally, feature selection methods helped improve the models' interpretability, making them useful for clinical applications. The study found that ML algorithms could complement traditional diagnostic approaches by reducing diagnostic errors and providing more objective assessments (Kumar & Das, 2021). Finnaly, Kumar and Das (2021) concluded that ML-based ASD diagnostic systems have the potential to enhance early detection and improve clinical decision-making. They recommended further research to refine these models and integrate them into healthcare systems. The authors emphasized the need for large-scale validation studies to ensure the generalizability of their findings and to enhance the practical application of AI-driven diagnostic tools (Kumar & Das, 2021).

**3.3 Deep Learning (DL) Models**

According to the study by Mukherjee et al. (2023), the primary objective is to review various machine learning models used for detecting autism spectrum disorder (ASD) and evaluate their effectiveness in clinical and non-clinical settings. The authors emphasize that early and accurate detection of ASD remains a significant challenge due to the heterogeneity of symptoms and the reliance on subjective assessments. The study systematically analyzes existing machine learning approaches, categorizing them based on supervised, unsupervised, and deep learning methodologies. The researchers evaluate different feature selection techniques, data sources, and classification algorithms, highlighting the strengths and weaknesses of each approach. They discuss how supervised learning models, particularly support vector machines and random forests, have been widely employed due to their interpretability and relatively high accuracy. However, deep learning models, including convolutional neural networks (CNNs) and recurrent neural networks (RNNs), have gained prominence for their ability to automatically extract complex patterns from behavioral and neuroimaging data. The study finds that while machine learning models show promise in ASD diagnosis, there are limitations related to dataset bias, lack of standardization, and the need for greater clinical validation. The authors conclude that future research should focus on integrating multi-modal data sources and refining models to enhance generalizability and clinical applicability (Mukherjee et al., 2023).

Also, the study by Sewani and Kashef (2020), explores the development of an autoencoder-based deep learning classifier for the efficient diagnosis of autism. The authors argue that traditional diagnostic methods, which rely on behavioral assessments and expert evaluations, are time-consuming and prone to subjectivity. Their study proposes an unsupervised learning model utilizing an autoencoder, a neural network architecture designed to learn efficient representations of input data without explicit labels. The methodology involves training the autoencoder on a dataset containing behavioral and neurophysiological features, allowing the model to learn the intrinsic structure of ASD-related characteristics. The classifier is then evaluated using multiple performance metrics, including precision, recall, and F1-score. The results indicate that the proposed model achieves higher classification accuracy compared to traditional machine learning techniques, demonstrating its potential as a reliable tool for ASD screening. The authors discuss the advantages of autoencoders in capturing non-linear relationships within the data, making them particularly useful for conditions like ASD, where symptoms manifest in diverse ways. However, they also acknowledge certain limitations, such as the need for large and well-annotated datasets, the risk of overfitting, and the challenge of interpretability in deep learning models. The study concludes that while autoencoder-based models represent a step forward in ASD diagnosis, further validation and clinical trials are necessary before widespread adoption (Sewani & Kashef, 2020).

Furthermore, Niu et al. (2020), examine the application of multichannel deep attention neural networks for classifying ASD using neuroimaging and personal characteristic data. The study is based on the premise that combining multiple data sources can improve diagnostic accuracy and provide deeper insights into ASD-related biomarkers. The researchers develop a deep learning model incorporating attention mechanisms, which allow the model to focus on the most relevant features in neuroimaging and behavioral data. Their methodology involves preprocessing neuroimaging scans to extract key features, which are then combined with demographic and behavioral information to train the deep learning model. The results demonstrate that the proposed model outperforms traditional classification methods by effectively identifying ASD-related patterns. The inclusion of an attention mechanism enhances interpretability, enabling researchers to understand which brain regions and behavioral traits contribute most to the diagnosis. However, the authors highlight several challenges, including the computational complexity of deep learning models, the difficulty of acquiring high-quality neuroimaging data, and ethical concerns regarding data privacy. They conclude that future research should aim to refine these models by incorporating larger and more diverse datasets, improving model efficiency, and exploring real-time applications in clinical settings (Niu et al., 2020).

A comparative analysis of these studies reveals significant progress in leveraging machine learning for ASD diagnosis. Mukherjee et al. (2023), provide a broad review of existing methodologies, emphasizing the need for integration and standardization in machine learning-based ASD detection. Sewani and Kashef (2020),contribute by demonstrating the potential of autoencoder-based models in improving diagnostic efficiency, particularly in cases where labeled data is scarce. Niu et al. (2020),further advance the field by incorporating neuroimaging data and attention mechanisms, highlighting the potential for multi-modal approaches in ASD classification. Despite these advancements, common challenges persist across all three studies, including dataset limitations, model interpretability, and the necessity for clinical validation. Future research should prioritize the development of standardized machine learning frameworks that integrate diverse data sources while ensuring model transparency and real-world applicability.

Furthermore, according to the study by Alsaidi et al. (2023), an approach is proposed for predicting Autism Spectrum Disorder (ASD) using convolutional neural networks (CNNs) and eye-tracking data. The research focuses on the potential of eye-tracking scan paths to serve as a diagnostic marker, allowing for an objective method of ASD assessment. The authors state that existing diagnostic methods rely on clinical evaluations and observations, which can lead to delayed or incorrect diagnoses (Alsaidi et al., 2023). Based on this framework, the study explores whether the use of deep learning techniques can enhance the accuracy and objectivity of diagnosis.The methodology used in the study includes collecting eye-tracking data from participants with and without ASD. These data are then preprocessed and fed into a CNN for training and classification. The authors describe the structure of the neural network, the data normalization processes, and techniques for preventing overfitting (Alsaidi et al., 2023). Particular emphasis is placed on the selection of hyperparameters and the use of cross-validation to ensure the generalizability of the results.

In the results, the study presents high classification accuracy rates, indicating that the proposed approach could serve as a useful diagnostic tool. Performance metrics of the model, such as accuracy, sensitivity, and specificity, demonstrate that eye-tracking data can reliably differentiate individuals with ASD from those without the disorder (Alsaidi et al., 2023). The authors also highlight that the effectiveness of their model compares favorably with other diagnostic methods while offering the advantage of automation.

In the discussion and conclusions, the study acknowledges certain limitations, such as the relatively small sample size and the need for further validation of results in larger populations. Despite these limitations, the authors emphasize the importance of using artificial intelligence in clinical applications and suggest further exploration of the combined use of eye-tracking and other biological markers to improve ASD diagnosis (Alsaidi et al., 2023).

## 4. aConclusions

This study examined the role of Artificial Intelligence (AI) in the diagnosis and intervention for individuals with Autism Spectrum Disorder (ASD), with a particular focus on the potential offered by modern technologies as well as the challenges they present when implemented in clinical and educational settings. The findings of the research underline the transformative possibilities of AI in improving the detection and management of ASD, positioning it as an effective and complementary tool for professionals, caregivers, and researchers in the field.

One of the primary conclusions drawn from the study is the significant contribution of AI to the early and more precise diagnosis of autism. By leveraging advanced machine learning (ML) and deep learning (DL) algorithms, AI can identify patterns in large datasets that are often difficult to detect through traditional diagnostic methods, such as clinical assessments and behavioral observations. Specifically, AI techniques can process and analyze multimodal data, including biometric information (such as heart rate, facial expressions, and body movement), linguistic patterns (such as speech delays, tone, and vocabulary), and non-verbal behaviors (such as gestures, eye contact, and social interactions). By utilizing neural networks and DL models, AI can provide highly accurate and objective diagnostic results that minimize human error and subjectivity. This capability is particularly valuable in the context of early diagnosis, as it facilitates quicker identification of ASD, which in turn allows for earlier intervention and support, ultimately improving the social, cognitive, and emotional development of children.

Beyond diagnosis, AI also holds great promise in advancing personalized and targeted interventions for individuals with ASD. Educational robots, such as NAO and Kaspar, have demonstrated positive effects in improving social skills and communication among children with autism. These robots are programmed to engage in structured, repetitive, and predictable interactions, which are essential for individuals with ASD, who may struggle with social cues and spontaneity. The robots' ability to provide consistent and adaptive responses helps children practice and internalize social behaviors in a safe and controlled environment. Moreover, AI-powered Augmentative and Alternative Communication (AAC) systems are enhancing the ability of non-verbal individuals with ASD to communicate more effectively. These systems utilize speech-generating devices or visual interfaces to enable users to express their needs, thoughts, and emotions in a way that would otherwise be difficult or impossible. Additionally, machine learning-based chatbots can facilitate the development of language skills by providing interactive dialogues tailored to the individual's communication level and needs.

However, despite the immense potential of AI in the field of ASD diagnosis and intervention, there are several challenges that must be addressed to fully realize its benefits. One of the major concerns is the limited long-term evaluation of AI-based interventions. Most existing studies focus on short-term outcomes, such as immediate improvements in social skills or communication, but there is still insufficient data on the long-term effectiveness and sustainability of these technologies in supporting individuals with ASD over time. Furthermore, the adaptation of AI technologies to meet the unique and individualized needs of each child remains an ongoing challenge. Unlike traditional therapeutic approaches that can be adjusted by clinicians based on their professional judgment, AI interventions must be highly customizable to account for the wide range of symptoms and abilities seen in individuals with ASD. To date, no universal model exists that can provide one-size-fits-all solutions, making it necessary for researchers to develop more flexible and adaptive AI systems.

Ethical and practical concerns also arise when implementing AI in this sensitive domain. Data privacy and security are paramount, as AI systems often rely on large volumes of personal and sensitive data, including biometric information and behavioral observations. Safeguarding this data to prevent misuse or breaches is a crucial aspect that needs to be addressed. Additionally, professionals who use AI-based tools must receive adequate training and support to ensure they can operate these technologies effectively and ethically. There is also a need for clear ethical guidelines that govern the development and deployment of AI in the autism community, particularly with respect to issues such as informed consent and the potential for AI systems to inadvertently reinforce biases or stereotypes.

In conclusion, AI presents an innovative and promising tool that has the potential to revolutionize the way autism is diagnosed, assessed, and treated. It can improve the accuracy and speed of diagnosis, enhance the personalization of interventions, and provide novel solutions for supporting the communication and social development of individuals with ASD. However, for AI to be successfully integrated into therapeutic and educational frameworks, further research is necessary to evaluate its long-term effectiveness, address ethical concerns, and create guidelines for its responsible use. The future challenge lies in the development of sustainable, scalable, and adaptable AI applications that not only enhance the quality of life for individuals with ASD but also ensure their accessibility and social acceptability across diverse cultural and societal contexts.